\documentclass{cCOS2e}

\usepackage{epstopdf}
\usepackage{colortbl}

\usepackage{subfig}
\usepackage{amsmath}
\usepackage[longnamesfirst,sort]{natbib}
\bibpunct[, ]{(}{)}{;}{a}{,}{,}

\theoremstyle{plain}

\theoremstyle{definition}

\theoremstyle{remark}

\begin{document}



\title{Improving interactive reinforcement learning:\\What makes a good teacher?}

\author{
\name{Francisco Cruz\textsuperscript{a,b,1}\thanks{$^1$Corresponding author. Email: cruz@informatik.uni-hamburg.de, francisco.cruz@ucentral.cl},
Sven Magg\textsuperscript{a}, Yukie Nagai\textsuperscript{c,2}\thanks{$^2$Yukie Nagai has been working at National Institute of Information and Communications Technology since May 2017.},
and Stefan Wermter\textsuperscript{a}}
\affil{\textsuperscript{a}Knowledge Technology Group, Department of Informatics, University of Hamburg, Germany.
\textsuperscript{b}Escuela de Computaci\'on e Inform\'atica, Facultad de Ingenier\'ia, Universidad Central de Chile.
\textsuperscript{c}Emergent Robotics Laboratory, Graduate School of Engineering, Osaka University, Japan.}
\received{v4.0 released April 2015}
}

\maketitle

\begin{abstract}
Interactive reinforcement learning has become an important apprenticeship approach to speed up convergence in classic reinforcement learning problems. 
In this regard, a variant of interactive reinforcement learning is policy shaping which uses a parent-like trainer to propose the next action to be performed and by doing so reduces the search space by advice. 
On some occasions, the trainer may be another artificial agent which in turn was trained using reinforcement learning methods to afterward becoming an advisor for other learner-agents. 
In this work, we analyze internal representations and characteristics of artificial agents to determine which agent may outperform others to become a better trainer-agent.
Using a polymath agent, as compared to a specialist agent, an advisor leads to a larger reward and faster convergence of the reward signal and also to a more stable behavior in terms of the state visit frequency of the learner-agents.
Moreover, we analyze system interaction parameters in order to determine how influential they are in the apprenticeship process, where the consistency of feedback is much more relevant when dealing with different learner obedience parameters.
\end{abstract}


\section{Introduction}

Reinforcement learning (RL) \citep{Sutton98} is a behavior-based approach which allows an agent, either an infant or a robot, to learn a task by interacting with its environment and observing how the environment responds to the agent's actions.
RL has been shown in robotics \citep{Kober13, Kormushev13} and in infant studies \citep{Hammerer12, Deak14} to be successful in terms of acquiring new skills, mapping situations to actions \citep{Cangelosi15}.

To learn a task, an RL agent has to interact with its environment over time in order to collect enough knowledge about the intended task. 
Nevertheless, on some occasions, it is impractical to leave the agent to only learn autonomously, mainly due to time restrictions and therefore, we aim to find a way to accelerate the learning process for RL.

In domestic and natural environments, adaptive agent behavior is needed utilizing approaches used by humans and animals. 
Interactive reinforcement learning (IRL) allows to speed up the apprenticeship process by using a parent-like advisor to support the learning by delivering useful advice in selected episodes. 
This allows to reduce the search space and thus to learn the task faster in comparison to an agent exploring fully autonomously \citep{Suay11, Cruz15}. 
In this regard, the parent-like teacher guides the learning robot, enhancing its performance in the same manner as external caregivers may support infants in the accomplishment of a given task, with the provided support frequently decreasing over time. 
This teaching technique has become known as parental scaffolding \citep{Breazeal98, Ugur15}.

The parent-like teacher can be either a human user or another artificial agent. 
By using artificial agents as teachers, some properties have been studied so far such as different effects of delivering advice in different episodes and with different strategies during the learning process \citep{Torrey13, Taylor14} and effects of different probabilities and consistency of feedback \citep{Griffith13, Cruz14, Cruz16}. 
Nonetheless, to the best of our knowledge, there is no study so far about the implications of utilizing artificial teachers with different characteristics and different internal representations of the knowledge based on their previous experience. 
Moreover, the effects when the learner ignores some of the advice has also not been studied in artificial agent-agent interaction, although some insights are given in Griffiths' work using human-human interaction with a computational interface \citep{Griffiths12}.

In this paper, we study effects of agent-agent interaction in terms of achieved learning when parent-like teachers differ in essence and when learner agents vary in the way they incorporate the advice. 
We have seen differences in the performance which could lead to adaptive behavior in order to reduce interactive feedback between trainer and learner.

This paper is organized as follows: in the second section, we present background and related work about IRL from both neuroscience and computational points of view. 
The third section shows the proposed IRL scenario which has been previously used but is updated here to further integrate multi-modal advice from human teachers. 
In the fourth section, we present the experimental set-up and obtained results. 
Finally, the fifth section gives an overall discussion including main conclusions and future work.

\section{Interactive Reinforcement Learning}
Learning in humans and animals has been widely studied by neuroscience yielding a better understanding of how the brain can acquire new cognitive skills. 
We currently know that RL is associated with cognitive memory and decision-making in animals' and humans' brains in terms of how behavior is generated \citep{Niv09}. 
In general, computational neuroscience has interpreted data and used abstract and formal theories to help to understand about functions in the brain.

In this regard, RL is a method used to address optimal decision-making, attempting to maximize collected reward and minimize the punishment over time. 
It is a mechanism utilized by humans and in robotic agents. 
In developmental learning, it plays an important role since it allows infants to learn through exploration of the environment and connect experiences with pleasant feelings which are associated with higher levels of dopamine in the brain \citep{Wise78, Gershman15}.

RL is a plausible method to develop goal-directed action strategies. During an episode, an agent explores the state space within the environment selecting random actions which move the agent to a new state. Moreover, a reward signal is received after performing an action, which may encode a positive compensation or a negative punishment. 
Over time, the agent learns the value of the states in terms of future reward, or reward proximity, and how to get to states with higher values to reach the target by performing actions \citep{Weber08}. 

\begin{figure}
\centering
\includegraphics[width=0.9\linewidth]{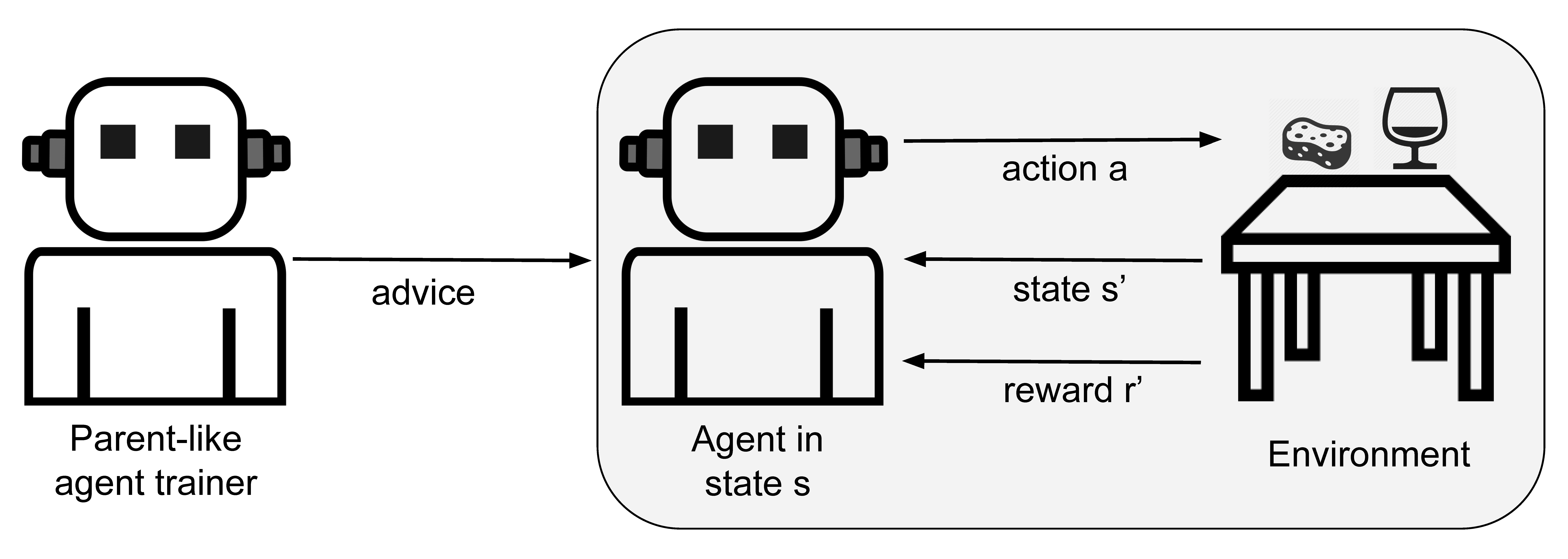}
\caption{An interactive reinforcement learning approach with policy shaping. The agent autonomously performs action $a$ in state $s$ obtaining reward $r'$ and reaching the next state $s'$. In selected states, the trainer advises the learner-agent changing the action to be performed in the environment.}
\label{fig:InterativeRLScenario}
\end{figure}

In robotics, RL has been used to allow robotic agents to autonomously explore their environment in order to develop new skills \citep{Wiering12, Mnih15}. 
To solve an RL problem means to find at least one optimal policy that collects the highest reward possible in the long run. 
Such a policy is known in psychology as a set of stimulus--response rules \citep{Kornblum90}. 
Optimal policies are denoted by $\pi^{*}$ and share the action-value function which is denoted by $q^{*}$ and defined as: $q^{*}(s,a) = \underset{\pi} {\mathrm{max}} ~q^{\pi}(s, a)$. 
The optimal action--value function can be solved through the Bellman optimality equation for $q^{*}$:

\begin{equation}
  q^{*}(s,a) = \sum_{s'} p(s'|s,a) [ r(s,a,s')+\gamma\max_{a'} q^{*}(s',a') ]
  \label{eq:bellman}
\end{equation}
where $s$ is the current state, $a$ is the taken action, $s'$ is the next state reached by performing the action $a$ in the state $s$, and $a'$ are possible actions that could be taken in $s'$. 
In the equation, $p$ represents the probability of reaching the state $s'$ given that the current state is $s$ and the selected action is $a$, and $r$ is the received reward for performing action $a$ in the state $s$ for reaching the state $s'$. 
The parameter $\gamma$ is known as discount rate and represents how influential future rewards are \citep{Sutton98}. 
The gray box in Fig. \ref{fig:InterativeRLScenario} shows the general description of the RL framework, where the environment is represented by domestic objects which are related to our scenario which is described in the next section.

In the learning phase, to solve equation \ref{eq:bellman}, one strategy is to allow the agent to perform actions considering transitions from state--action pair to state--action pair rather than transitions from state to state only. 
Accordingly, the on-policy method SARSA \citep{Rummery94} updates every state--action value according to the equation:

\begin{equation}
Q(s,a) \leftarrow Q(s,a) + \alpha[r' + \gamma Q(s',a') - Q(s,a)]
\label{eq:SARSA}
\end{equation} 
where $Q$ is the value of the state--action pair and $\alpha$ the learning rate.

Although the next action can be autonomously selected by choosing the best known action at the moment, represented by the highest state--action pair, an intuitive strategy to speed up the learning process would be to include external advice in the apprenticeship loop; early research on this topic using both humans and robots can be found in \citep{Lin91}. 
When using IRL, an action is interactively encouraged by a trainer with a priori knowledge about the desired goal \citep{Thomaz05, Thomaz06, Knox13}. 
In IRL, using a trainer to advise an agent on future actions is known as policy shaping \citep{Cederborg15, Amir16}.

Supportive advice can be obtained from diverse sources like expert and non-expert humans, artificial agents with perfect knowledge about the task, or previously trained artificial agents with certain knowledge about the task. 
In this work, an artificial trainer-agent which was itself previously trained through RL is used to provide advice, which has been formerly used in other works. 
For instance, in \citep{Cruz14} advice is given based on an interaction probability and consistency of feedback. In Taylor's works, interaction is based on a maximal budget of advice and they studied which moment is better to give advice during the training \citep{Torrey13, Taylor14}. 
Fig. \ref{fig:InterativeRLScenario} shows a general overview of the agent--agent scheme where the trainer provides advice in selected episodes to the learner-agent to bootstrap its learning process.

Although interactive advice improves the learning performance of learner-agents, a problem which remains open and that can significantly affect the agent's performance is the need of a good trainer since consecutive mistakes may lead to a worse training time \citep{Cruz16}.
In principle, one may think that an expert agent with a larger accumulated reward should be a good candidate to become the trainer.
Expert agents, either human or artificial, have been used in different reinforcement learning approaches using advice (e.g.: \cite{daSilva17, Ahmadabadi02, Ahmadabadi00, Prince99}).
However, when we look into the internal knowledge representation, this may not necessarily be the best option.
On some occasions, agents with lower overall performance may be better trainers due to a possibly vast experience about less common states (i.e. states that do not necessarily lead to the optimal performance) and therefore, may give better advice in those states.
Some insights on using trainer-agents with different abilities have been discussed by \cite{Taylor11} in a simulated robot soccer domain by using a human-agent transfer approach.

\section{Domestic Robot Scenario}
In this paper, we extend a previously used RL scenario which consists of a robotic agent performing a cleaning task \citep{Cruz16}. 
Here, we do not deal with contextual affordances and, therefore, we do not have to previously learn them which results in a shorter training time, in general.

The current scenario comprises two objects, three locations, and seven actions. 
The robot is placed in front of a table in order to clean it up. 
In this scenario, there are two objects: a \textit{cup} which is initially at a random location on the table and needs to be relocated as the table is being wiped, and a \textit{sponge} which is used by the robot in order to wipe different sections of the table.

Three locations have been defined in the cleaning scenario: \textit{left} and \textit{right} to refer to each of the two sections of the table, and one additional position called \textit{home} which is the robot's arm's initial position and the location where the sponge is placed when not being used. 
Furthermore, seven domain-specific actions are allowed in this scenario defined as follows:

\begin{enumerate}[\hspace{0cm}(P1)]	
	\item[(i)] \texttt{GET:} allows the robot to pick up the object which is placed in the same location as its hand.
	\item[(ii)] \texttt{DROP:} allows the robot to put down the object held in its hand. The object is placed in the same location where the hand is.
	\item[(iii)] \texttt{GO HOME:} moves the hand to the home position.
	\item[(iv)] \texttt{GO LEFT:} moves the hand to the left position.
	\item[(v)] \texttt{GO RIGHT:} moves the hand to the right position.
	\item[(vi)] \texttt{CLEAN:} allows the robot to clean the section of the table at the current hand position if holding the sponge.
	\item[(vii)] \texttt{ABORT:} cancels the execution of the cleaning task at any time and returns to the initial state.
\end{enumerate}

Each state is represented by using a state vector of four variables:

\begin{enumerate}[\hspace{0cm}(P1)]	
	\item[(i)] the object held in the agent's hand (if any), 
	\item[(ii)] the agent's hand position, 
	\item[(iii)] the position of the cup, and 
	\item[(iv)] a 2-tuple with the condition of each side of the table, i.e. whether the table surface is clean or dirty.
\end{enumerate}
	
Therefore, the state vector at any time $t$ is characterized as follows:

\begin{equation}
s_t = <handObject, handPosition, cupPosition, sideCondition>.
\end{equation}

\begin{figure}
\centering
\includegraphics[width=0.7\linewidth]{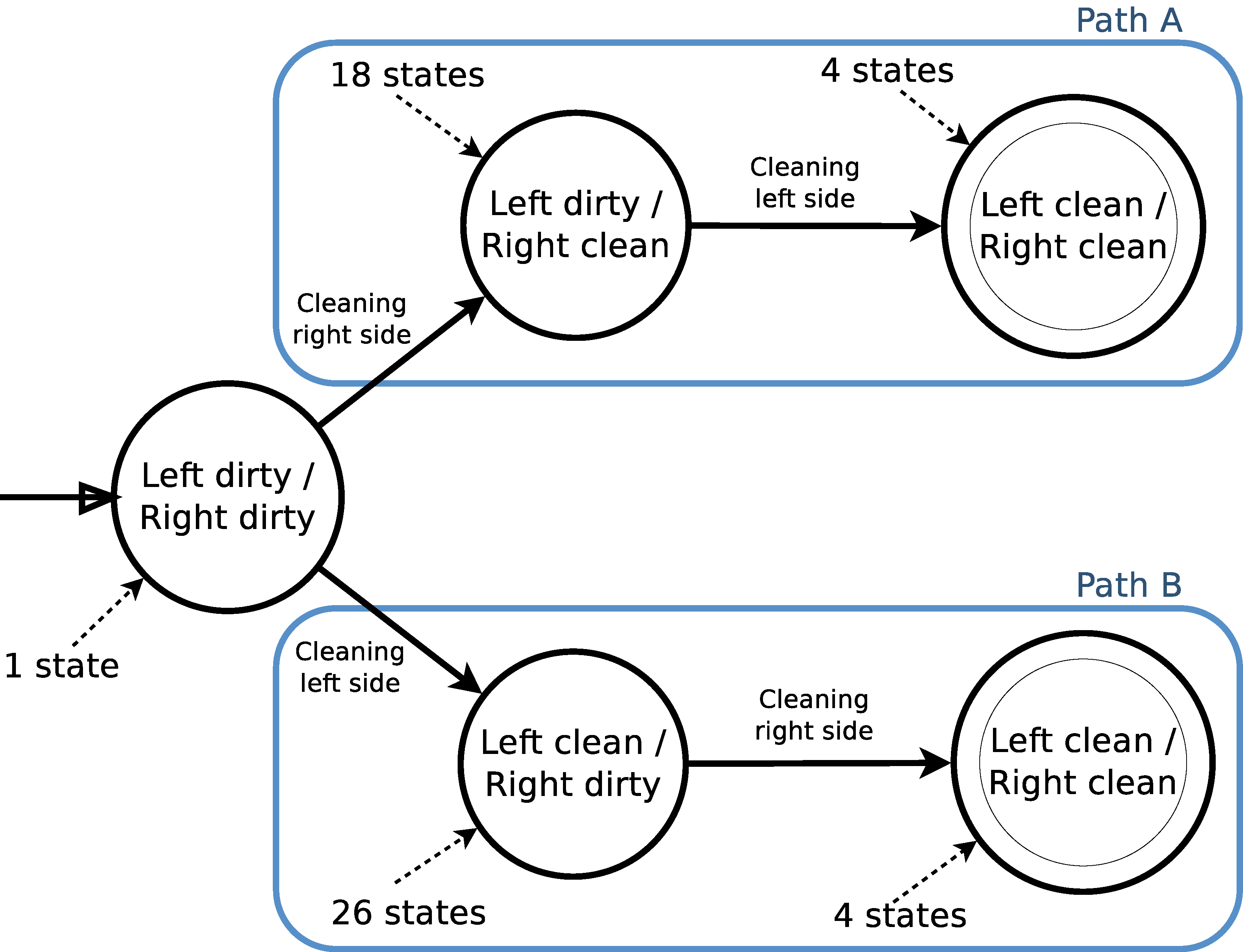}
\caption{Outline of state transitions in the defined cleaning scenario. Two different paths are possible to reach a final state. Each path implies a different number of intermediate states which influence the total amount of collected reward during a learning episode. Thus path A comprises 23 states and path B 31 states.}
\label{fig:extendedSolution}
\end{figure}

As long as the agent successfully finishes the task, a reward equal to $1$ is given to it, whereas a reward of $-1$ is given if a failed-state was reached. 
In this context, a failed-state is a state from where the robot cannot continue the expected task execution, for instance attempting to pick-up an object when it is already holding another object. 
Furthermore, it is given a small negative reward of $-0.01$ to encourage the agent to take shorter paths towards a final state. 
Therefore, the reward function can be posed as:

\begin{equation}
r(s) = \left\{
\begin{array}{r l}
  1 & \textrm{if } s \textrm{ is the final state}\\
 -1 & \textrm{if } s \textrm{ is a failed-state}\\
 -0.01 & \textrm{otherwise}
\end{array}
\right.
\label{eq:reward}
\end{equation}

At the beginning of each training episode, the robot's hand is free at the home location, the sponge is also placed at the home position, while the cup is at either the left or the right location, and both table sections are dirty. 
Therefore, the initial state $s_0$ may be represented as:

\begin{equation}
s_0 = <\mathit{free, home, left | right, [dirty, dirty]}>.
\end{equation}

From the initial state, the state vector is updated every time after performing an action according to the state transition table as shown in Table \ref{tb:vectorUpdate}. 
In the current scenario, considering the state vector features, there are 53 different states which represent two divergent paths to two final states. 
Fig. \ref{fig:extendedSolution} depicts a summarized illustration of the transitions to reach a final state assuming the cup to be initially at the left position. 
The figure also shows the number of states involved in each path. 
Therefore, each path leads to a different number of transited states which in turn also leads to a different accumulated reward.

\begin{table}
\tbl{State vector transitions. After performing an action the agent reaches either a new state or a failed condition, if the latter, the agent starts another training episode from the initial state $s_0$.}
{\begin{tabular}[l]{@{}ll}\toprule
  \textbf{Action} & \textbf{State vector update} \\
\colrule
  Get & \textbf{if} handPos == \textit{home} \textbf{\&\&} handObj == \textit{cup} \textbf{then} FAILED \\
      & \textbf{if} handPos == \textit{cupPos} \textbf{\&\&} handObj == \textit{sponge} \textbf{then} FAILED \\
      & \textbf{if} handPos == \textit{home} \textbf{then} handObj = \textit{sponge} \\
      & \textbf{if} handPos == cupPos \textbf{then} handObj = \textit{cup} \\
\colrule   
  Drop & \textbf{if} handPos == \textit{home} \textbf{\&\&} handObj == \textit{cup} \textbf{then} FAILED \\
       & \textbf{if} handPos != \textit{home} \textbf{\&\&} handObj == \textit{sponge} \textbf{then} FAILED \\
       & \textbf{otherwise} handObj = \textit{free} \\
\colrule
  Go $<$pos$>^{\rm *}$ & handPos = pos \\
               & \textbf{if} handObj == \textit{cup} \textbf{then} cupPos = pos \\
\colrule
  Clean & \textbf{if} handPos == cupPos \textbf{then} FAILED \\
        & \textbf{if} handPos == \textit{home} \textbf{then} FAILED \\
        & \textbf{if} handObj == \textit{sponge} \textbf{then} sideCond[handPos] = \textit{clean} \\
\colrule
  Abort & handPos = \textit{home} \\
        & handObj = \textit{free} \\
        & cupPos = random(pos) \\
        & sideCond = [dirty]*$|$pos$|$ \\
\botrule
\end{tabular}}
\tabnote{$^{\rm *}$ $<$pos$>$ may be any defined location, therefore three actions are represented by this transition, i.e.: go left, go right, and go home.}
\label{tb:vectorUpdate}
\end{table}

As defined, the same transitions may be used in scaled-up scenarios where more locations are defined on the table in a larger grid since the definition of transitions is done by only considering the object held by the robot and the hand position in reference to either the home location or the cup position.

\begin{figure}
\centering
\includegraphics[width=0.75\linewidth]{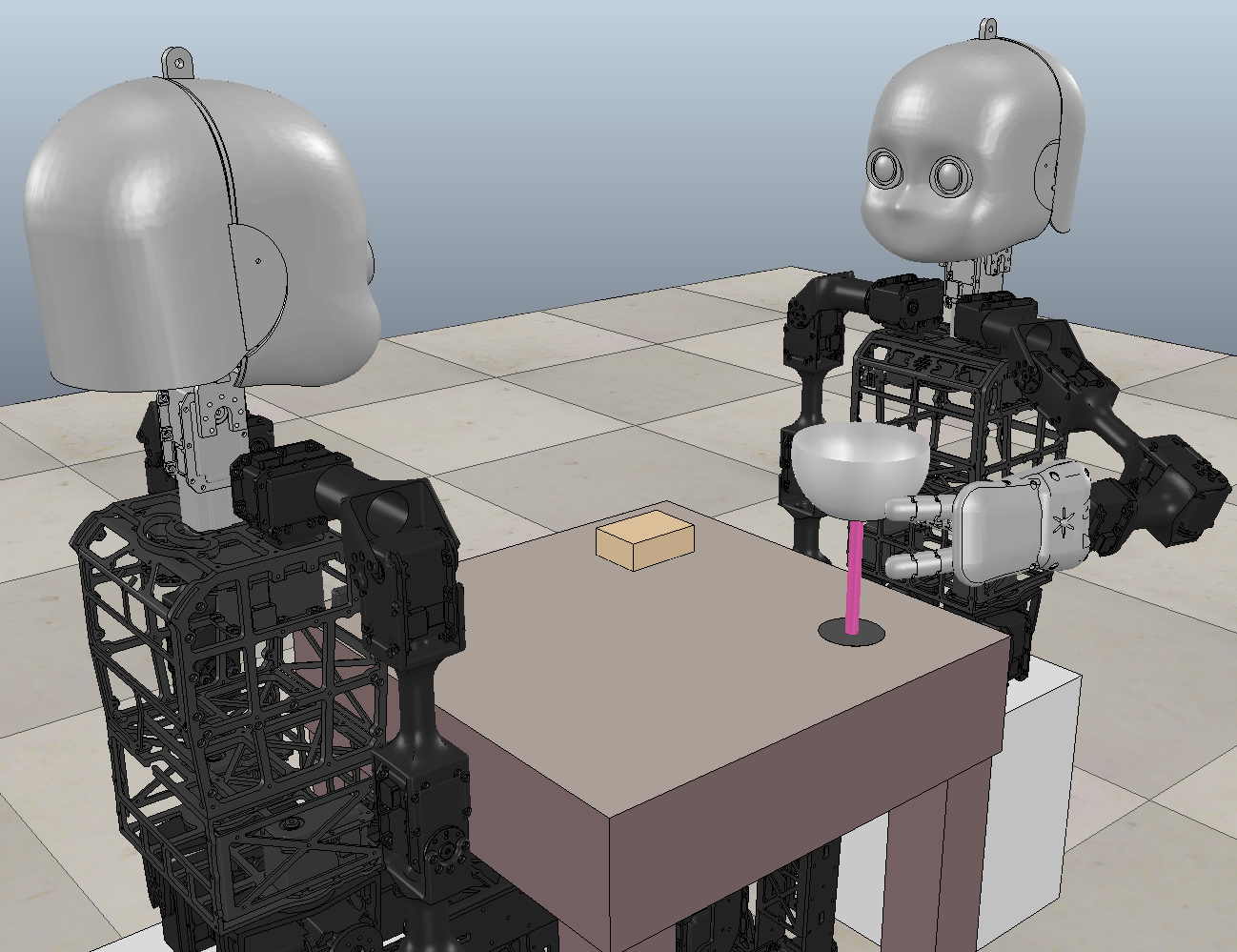}
\caption{Two robotic agents performing a domestic task in the defined home scenario. The trainer-agent advises the learner-agent in selected states what action to perform next.}
\label{fig:twoNICOs}
\end{figure}

Fig. \ref{fig:twoNICOs} shows the domestic robotic scenario with two robotic agents where one agent becomes the trainer by learning the task using autonomous RL. 
The second agent performs the same task supported by the trainer-agent with selected advice using the IRL framework.

\section{Experiments and Results}

In the following subsections, the experimental set-up will be explained in detail. 
Initially, we look into the internal representation and visited states of prospective advisor agents in order to explore which features may be important to act as a good trainer. 
Afterward, we compare the behavior of both the advisor and the learner in terms of the internal representation, visited states, and collected reward. 
Finally, we evaluate some system interaction parameters like frequency of feedback, consistency of feedback, and learner behavior.

All experiments included the training of $100$ agents through $3000$ episodes. 
Q-values were randomly initialized using a uniform distribution between $0$ and $1$. 
Other parameter values were learning rate $\alpha = 0.3$ and discount factor $\gamma = 0.9$.
Besides this, we used $\epsilon$-greedy action selection with $\epsilon = 0.1$. 
To assess the interaction between learner and trainer-agents we used a probability of feedback of $0.25$ as a base; nevertheless, we afterward varied this parameter along with the consistency of feedback and learner behavior. 
All the aforementioned parameters were empirically determined and related to our scenario.

\subsection{Choosing an Advisor Agent}

To acquire a sample of trainer-agents, autonomous RL was performed with $100$ agents, each of them a prospective trainer for the IRL approach. 
In the presented scenario, there are agents with diverse behaviors which differ mostly in the path they choose until reaching a final state. 
First, there are agents which most of the time choose the same path to complete the task, either path A or path B, which leads to a biased behavior due to the way the knowledge is acquired during the learning process. 
From this kind of behavior and taking into account our scenario, there exist agents that regularly take the shorter path (path A) and others that take the longer one (path B); we refer to them as the specialist-A and the specialist-B agents respectively. 
In both cases, agents successfully accomplish the task, although they accumulate different amounts of average reward.
Obviously, the specialist-A agents are the ones with better performance in terms of collected reward since fewer state transitions are needed to reach the final state. 
Second, there are agents with a more homogeneously distributed experience, meaning that they do not have a favorite sequence to follow and have equally explored both paths. 
We refer to such agents as polymath agents.

\begin{figure}
\centering
\makebox[\linewidth][c]{\includegraphics[width=1.0\linewidth]{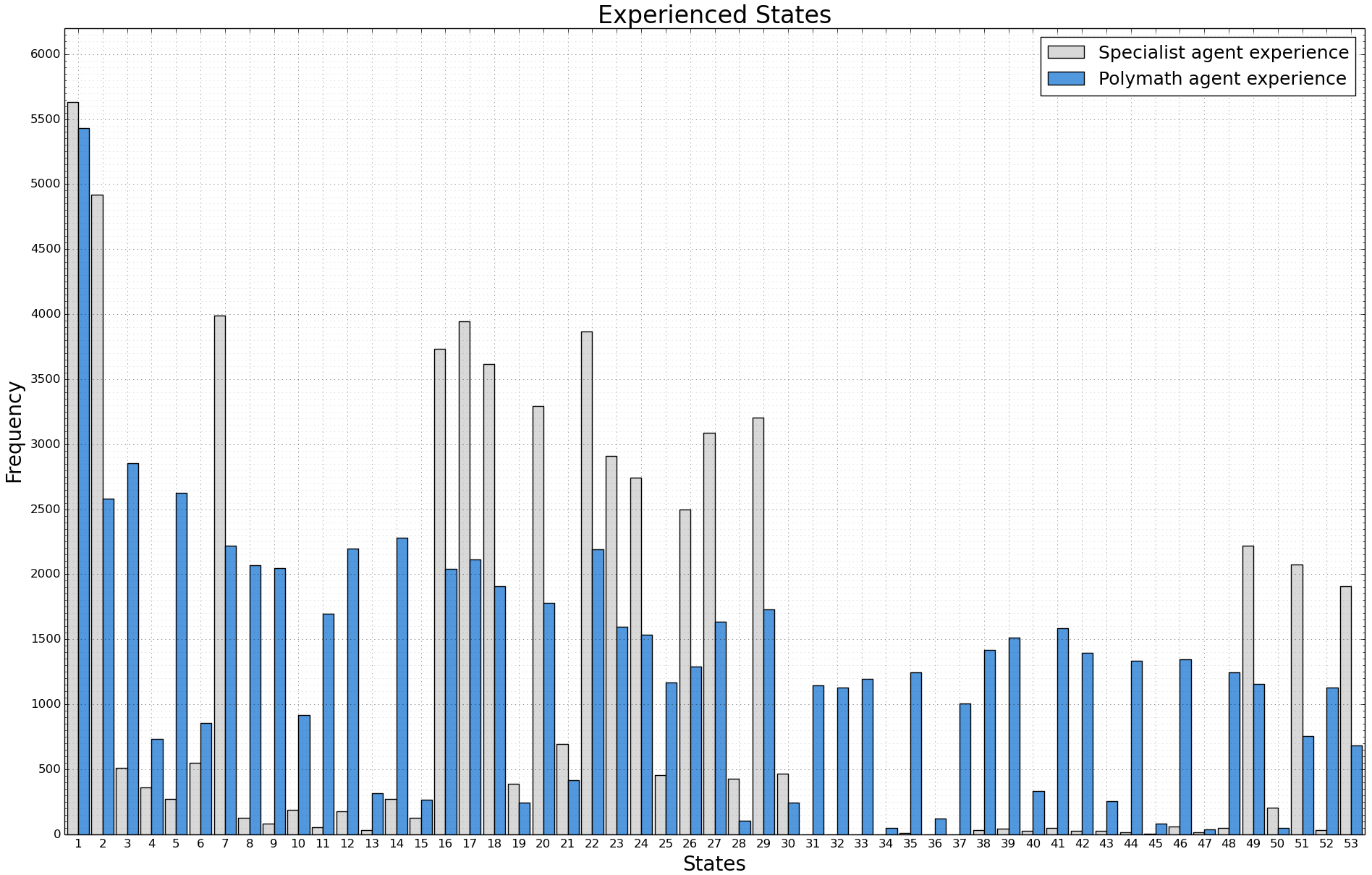}}
\caption{Frequency of visits per state for two agents. It is possible to observe two different behaviors. The biased (specialist-A) agent gained experience mostly on the shorter path, whereas the homogeneously-distributed (polymath) agent gained experience through most states.}
\label{fig:visitedTwoTeachers}
\end{figure}

To illustrate this, Fig. \ref{fig:visitedTwoTeachers} shows a frequency histogram of visited states for two potential trainer-agents over all training episodes. 
The histogram shows two distinct distributions, one for a specialist-A agent in gray and one for a polymath agent in blue. 
The specialist-A agent decided to clean the table following the shorter path most of the time and, therefore, there is an important concentration of visits among the states from $16$ to $29$ which are intermediate states to complete the task on this path.
Furthermore, there is a clear subset of states which was never visited during the learning. 
In contrast, the polymath agent visited all the states and transits on both paths to a similar extent.
In the case of the specialist-B agent, there is also a concentration of visits among a subset of states, similarly to the specialist-A agent. The specialist-B agent
decided most of the time to clean following the longer path along the states from $30$ to $48$ and barely visiting states from $16$ to $29$. Therefore, we do include this agent in the results hereafter but we do not present it in some plots to make the relevant information more accessible.

To further analyze the agents' behavior we took three representative agents, one per class, that we will from now on use with the respective names: specialist-A agent with biased behavior for the shorter path, specialist-B agent with biased behavior for the longer path, and polymath agent with unbiased behavior. 
The specialist-A agent visited each state with an average of $s_1=1121.21$ times, a standard deviation of $\sigma_s^1=1570.75$, an accumulated average reward of $r_1=0.11105$ per episode, and $R_1=333.15$ during the whole training. 
The specialist-B agent visited each state on average $s_2=1561,15$ times obtaining a more diverse experience than the previous agent but certainly not homogeneously distributed, which can also be appreciated in the standard deviation of $\sigma_s^2=1628.70$. 
The specialist-B agent accumulated an average reward of $r_2=-0.17839$ for each episode and a total of $R_2=-535.18$. 
In the case of the polymath agent, each state was visited an average of $s_3=1307.51$ times with standard deviation of $\sigma_s^3=947.96$. 
The accumulated average reward was $r_3=-0.00427$ per episode and the total reward was $R_3=-12.82$ during the whole training. 
Table \ref{tb:performance} shows a summary of the performance of the three aforementioned agents.

\begin{table}
\tbl{Visited states, standard deviation, reward accumulated per episode, and total collected reward for three agents from classes with different behavior. The agents show different characteristics as result of the autonomous learning process.}
{\begin{tabular}[l]{@{}lllllll}\toprule
  \textbf{Agent} & $\pmb{s}$ & $\pmb{\sigma_s}$ & $\pmb{r}$ & \textbf{R} & \textbf{Characteristic} \\
  
\colrule
  Specialist-A agent & 1121.21 & 1570.75 & 0.11105 & 333.15 & Largest accumulated reward \\
\colrule
  Specialist-B agent & 1561.15 & 1628.70 & -0.17839 & -535.18 & Largest amount of experience\\
\colrule
  Polymath agent & 1307.51 & 947.96 & -0.00427 & -12.82 & Smallest standard deviation \\

\botrule
\end{tabular}}

\label{tb:performance}
\end{table}

Nevertheless, accumulating plenty of reward does not necessarily lead to becoming a good trainer.
In fact, it only means that the agent is able to select the shorter path most of the time from the initial state, but the experience collected in other states not involved in that route is absent or barely present and therefore, such an agent cannot give good advice in those states where it does not know how to act optimally.

For a good trainer to emerge with knowledge of most of the situations or in all possible states we suggest an agent with a small standard deviation $\sigma_s$ from the mean frequency over all visited states, which represents a better distribution of the experience during the training. 
We select the trainer-agent $T^*$ computing:

\begin{equation}
  T^* = \underset{i \in A} {\mathrm{argmin}} ~\sigma_s^i
  \label{trainer}
\end{equation}
where $A$ is the set of all the trained agents and their respective visited states during the learning process.

Therefore, we propose that a good trainer is, in essence, an agent which not only collects more rewards but shows also a fairly distributed experience. 
From the three agents shown above, the polymath agent has a standard deviation of $\sigma_s=947.96$ and thus might be a good advisor. 
In Fig. \ref{fig:visitedTwoTeachers}, the experience distribution of such an agent is shown in blue and this experience distribution suggests that the agent has the knowledge to advise what action to perform in most of the states. 
In the case of the initial state, the frequency is much higher in comparison since this state is visited every time at the beginning of a learning episode. 
In fact, similar frequencies are observed in this state for a biased distribution.

\begin{figure}
\centering
\makebox[\linewidth][c]{\includegraphics[width=1.1\linewidth]{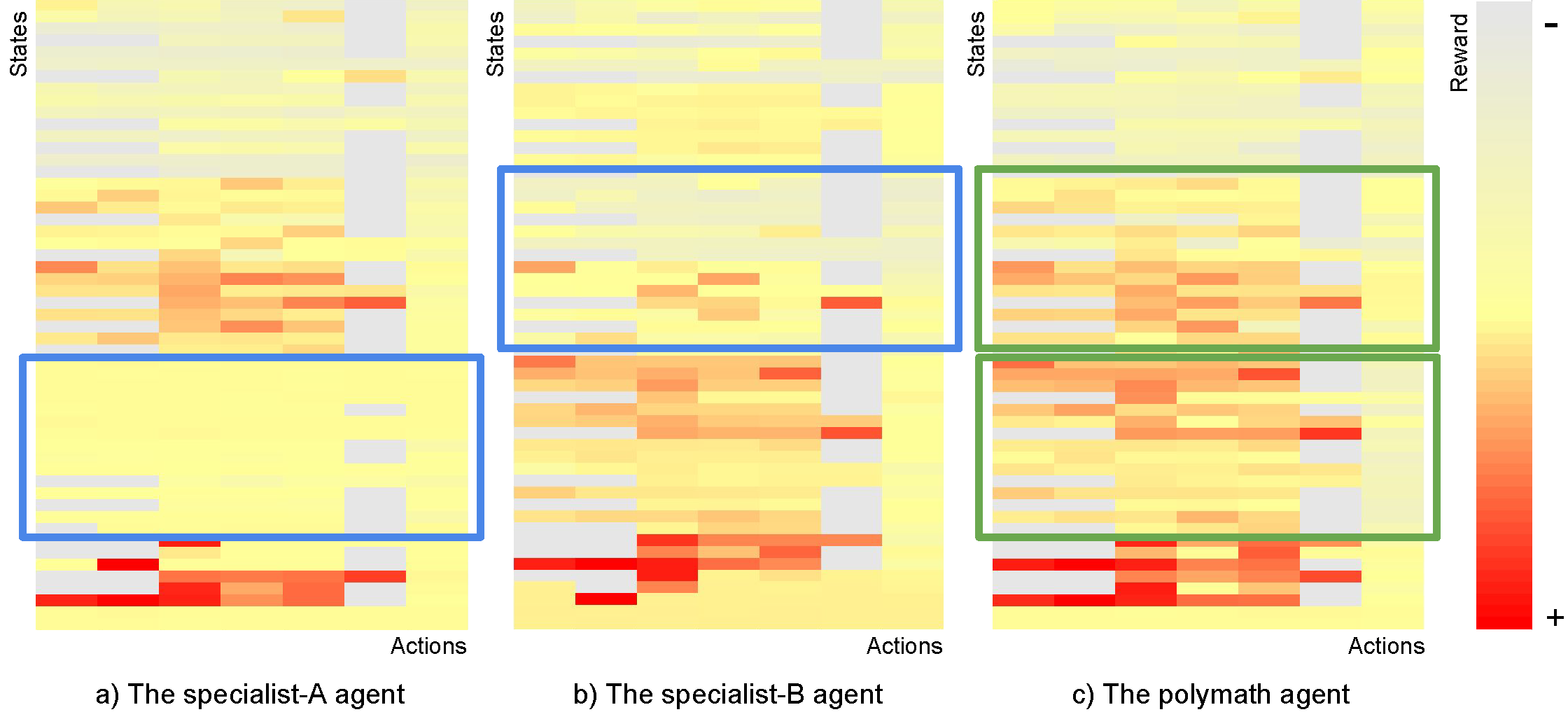}}
\caption{Internal knowledge representation for three possible parent-like advisors in terms of Q-values, namely the specialist-A, the specialist-B, and the polymath agent. 
The specialist-A agent shown in figure a), despite collecting more reward, does not have enough knowledge to advise a learner in every situation represented by the blue box. 
A similar situation is experienced by the specialist-B agent, as shown in figure b). 
The polymath agent shown in figure c) has overall much more distributed knowledge which allows it to better advise a learner-agent.}
\label{fig:QValues}
\end{figure}

We also recorded the internal representation of the knowledge through the Q-values to confirm the lack of learning in a subset of states.
Fig. \ref{fig:QValues} shows a heat map of the internal Q-values of three agents, the specialist-A, the specialist-B, and the polymath agent. 
Warmer regions represent a larger reward and colder regions lower values. 
In fact, the coldest regions are associated with failed-states from where the agent should start a new episode, obtaining a negative reward of $r=-1$ according to Eq. \ref{eq:reward}. 
In Fig. \ref{fig:QValues}, it can be observed that the specialist-A agent may be an inferior advisor since there exists a whole region uniformly in yellow, which shows no knowledge about what action to prefer. 
In the case of the specialist-B agent, there exists a region which shows much less knowledge on what action to prefer when comparing it with the two other agents.
In other words, the learned policies are partially incomplete as highlighted by the blue boxes in Fig. \ref{fig:QValues}.
To the contrary, the policy learned by the polymath agent is much more complete when observing the same regions as highlighted by the green boxes.
It is important to note that the region on top is in all cases colder than the rest because it is the most distant one from the final states where a positive reward $r=1$ is given, but in spite of that, the polymath agent is still able to select a suitable action according to the learned policy.

\subsection{Comparing Advisor and Learner Behavior}

Once we had chosen trainer-agents, we were able to compare how influential such a trainer was in the learning process of a learner. 
We used two agents shown in the previous subsection, the specialist-A and the polymath agent, the former with the largest accumulated reward and the latter with the smallest standard deviation. 

\begin{figure}
\centering
\makebox[\linewidth][c]{\includegraphics[width=1.0\linewidth]{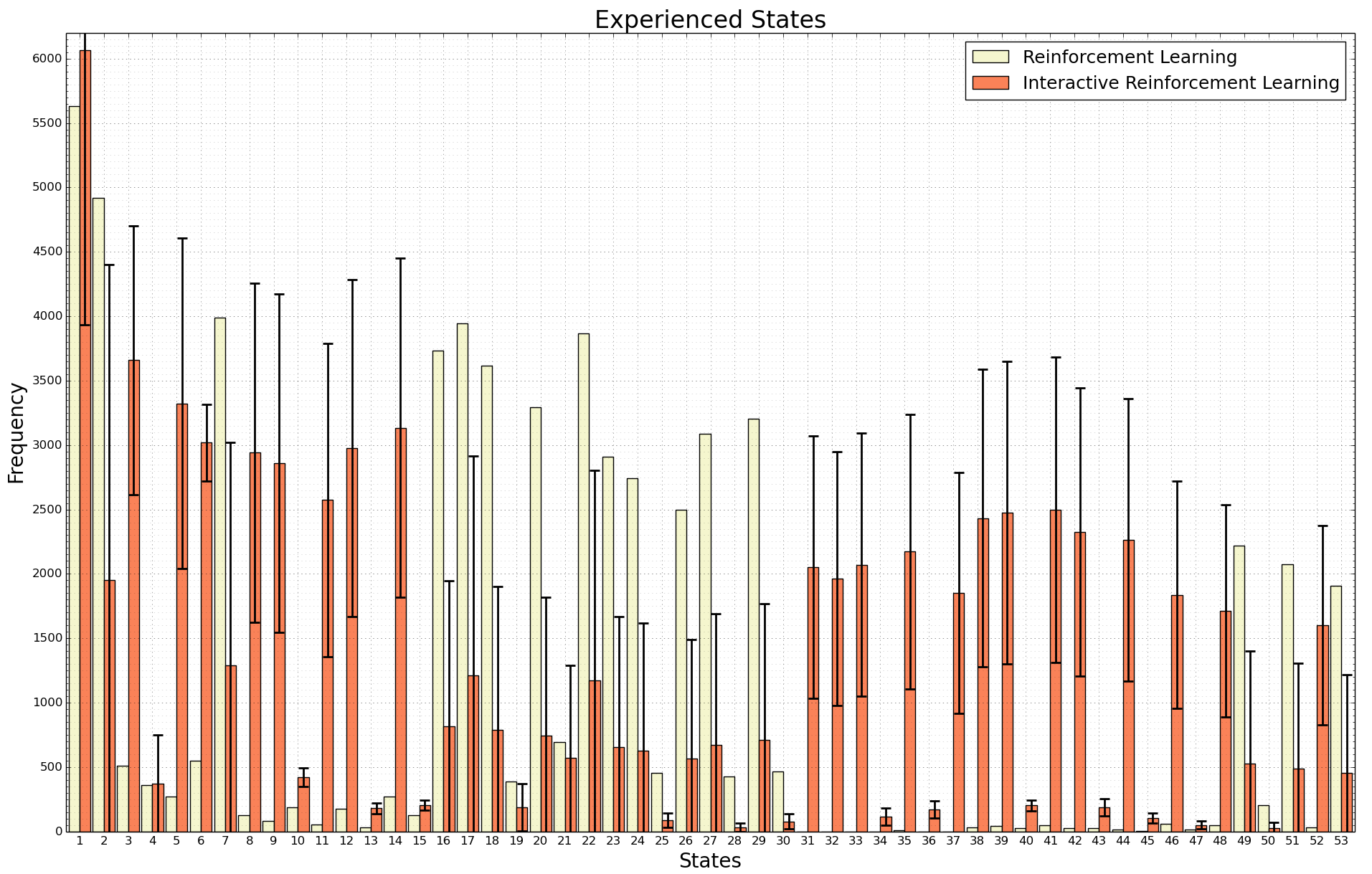}}
\caption{Visited states for the specialist-A RL trainer-agent and average state visits of IRL learner-agents. The averaged frequency for IRL agents moreover includes the standard deviation for visited states showing that in many cases the trainer-agent does not know how to advise and in consequence leads the learner-agent to dissimilar behavior.}
\label{fig:visitedRLvsIRLBadTeacher}
\end{figure}

\begin{figure}
\centering
\makebox[\linewidth][c]{\includegraphics[width=1.0\linewidth]{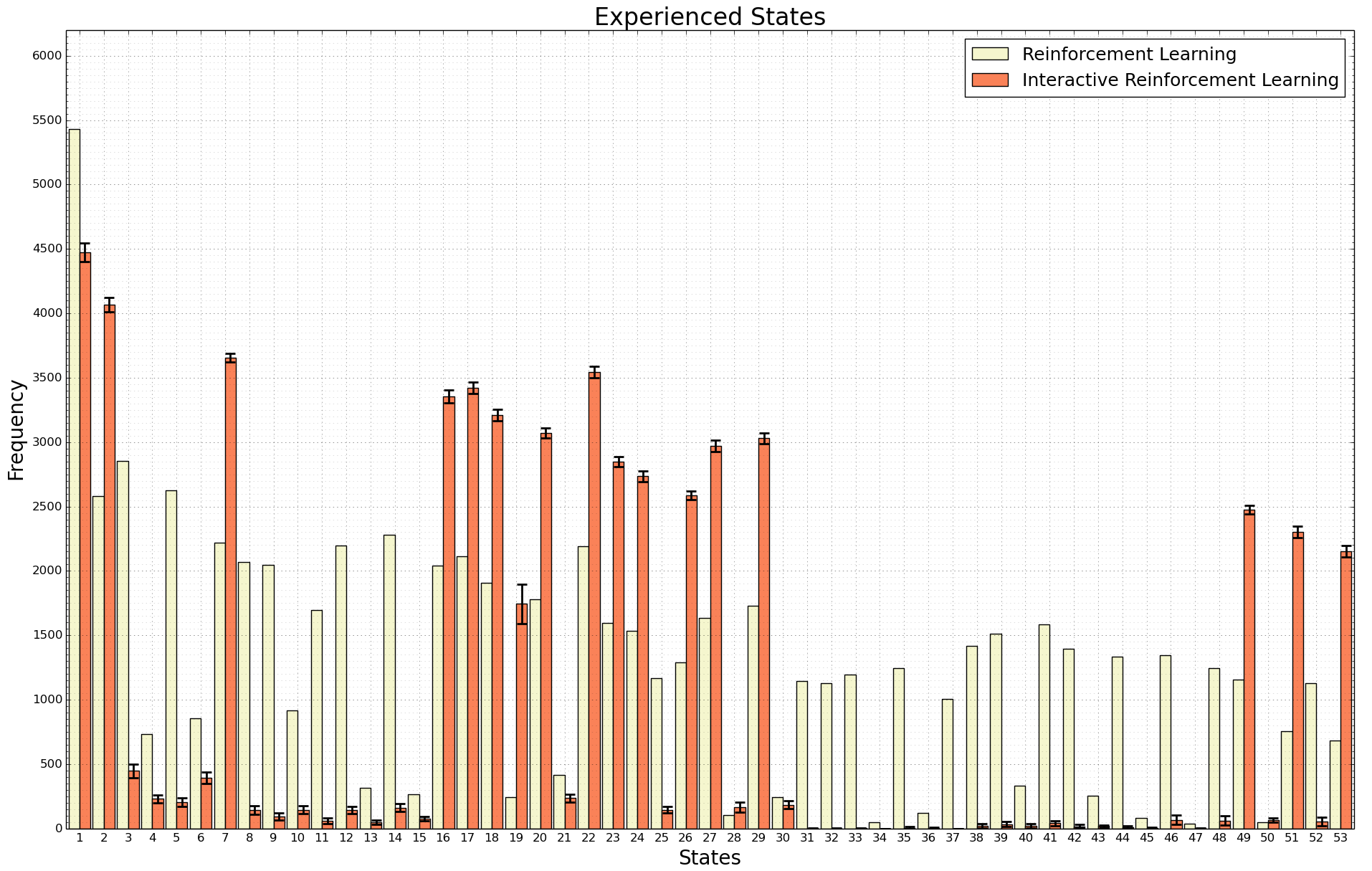}}
\caption{Visited states for the polymath RL trainer-agent and average state visits of IRL learner-agents. The averaged frequency for IRL agents includes the standard deviation which in this case is considerably lower as the learners are assisted by a trainer with more knowledge about the task-space which also leads learner-agents to have more stable behavior as they are consistently advised.}
\label{fig:visitedRLvsIRL}
\end{figure}

Fig. \ref{fig:visitedRLvsIRLBadTeacher} shows the frequency with which each state was visited for $100$ learner-agents on average using the specialist-A agent with biased frequency distribution as a trainer. 
We can observe a large standard deviation for visited states in IRL agents in most of the cases, which suggests diversity in terms of frequency for those states among the learner-agents. 
Fig. \ref{fig:visitedRLvsIRL} shows the average frequency of visits for each state for $100$ learner-agents using the polymath agent as a trainer which has a more homogeneous frequency distribution. 
It can be observed that the standard deviation for visited states in IRL agents is much lower in comparison to the previous case. 
This shows a more stable behavior in terms of visiting frequency in learner-agents when using the polymath trainer-agent.

\begin{figure}
\centering
\includegraphics[width=0.8\linewidth]{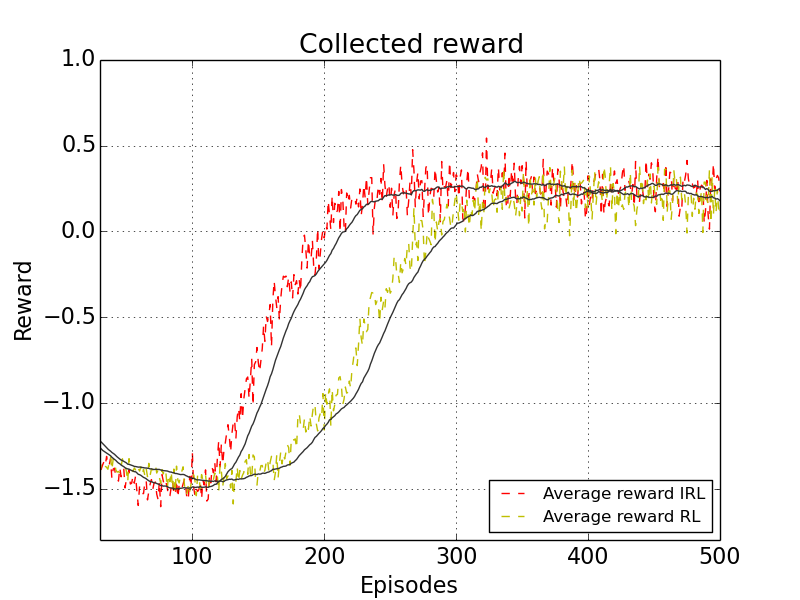}
\caption{Average collected reward by $100$ agents using RL and IRL approaches. In this case, a biased trainer (the specialist-A agent) is used to advise the learner-agents. The advice slightly improves the performance in terms of accumulated reward and convergence speed.}
\label{fig:rewardRLvsIRLBadTeacher}
\end{figure}

By using the specialist-A agent as a trainer in our IRL approach the average collected reward is slightly higher in comparison with autonomous RL.
In general, the IRL approach collects the reward faster than RL but in a similar magnitude after $400$ episodes. 
Fig. \ref{fig:rewardRLvsIRLBadTeacher} depicts the average collected reward during the first $500$ episodes using autonomous RL and IRL approaches with yellow and red respectively using the specialist-A agent as the trainer in the case of IRL. 
The gray curves show the convoluted collected reward inside a window of 30 values to smooth the results shown.

\begin{figure}
\centering
\includegraphics[width=0.8\linewidth]{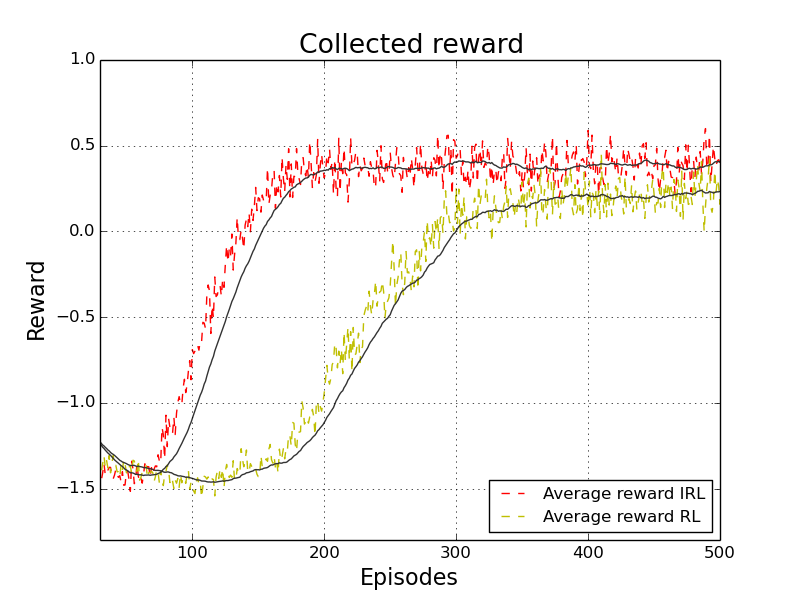}
\caption{Average collected reward by $100$ agents using RL and IRL approaches. When using an unbiased trainer-agent (the polymath agent), the accumulated reward is higher and the convergence speed faster in comparison with the previous case using a biased agent as an advisor.}
\label{fig:rewardRLvsIRL}
\end{figure}

On the other hand, by using the polymath agent as the trainer the IRL approach converges both faster and to a higher amount of reward when compared with the previous case. 
This is due to the polymath agent which knows the task-space better and is able to advise correctly in more situations than the specialist agent. 
In consequence, this allows the learner to complete the task faster and therefore accumulate more reward. Fig. \ref{fig:rewardRLvsIRL} shows the average collected reward in $500$ episodes for RL and IRL approaches. 
Once again, the gray curves show the convoluted collected reward inside a window of 30 values to smooth the results shown.
In the following experiments, only smooth curves will be used to simplify the analysis of the results.

Therefore, IRL is in general beneficial for a learner-agent in terms of accumulated reward and convergence speed. 
Nevertheless, the selection of the trainer can have significant implications on the learner's performance. 
In the following subsection, we analyze the main interaction parameters in order to understand how influential they are regarding the learner's performance when being advised by a potentially good trainer.

\subsection{Evaluating Interaction Parameters}

As part of this study, we evaluated the involved interaction parameters namely probability of feedback ($\mathcal{L}$), consistency of feedback ($\mathcal{C}$), and whether the learner follows the received advice or not in order to mimic actual human-human behavior where the learner occasionally
does not follow the advice \citep{Griffiths12}. 
We called this parameter \textit{learner obedience} $\mathcal{O} \in [0, 1]$, $0$ being an agent that never follows the advice and thus corresponds to a pure RL learner. Probability and consistency of feedback correspond to the frequency of giving advice to the learner and the degree to which such advice is rational in the current state respectively.

\begin{figure}
\centering
\subfloat{\includegraphics[width=0.5\textwidth]{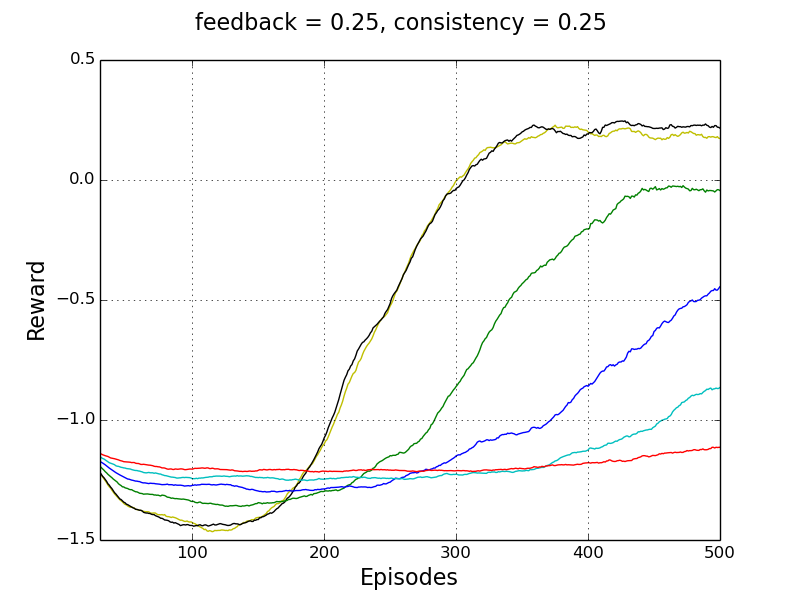}}
\subfloat{\includegraphics[width=0.5\textwidth]{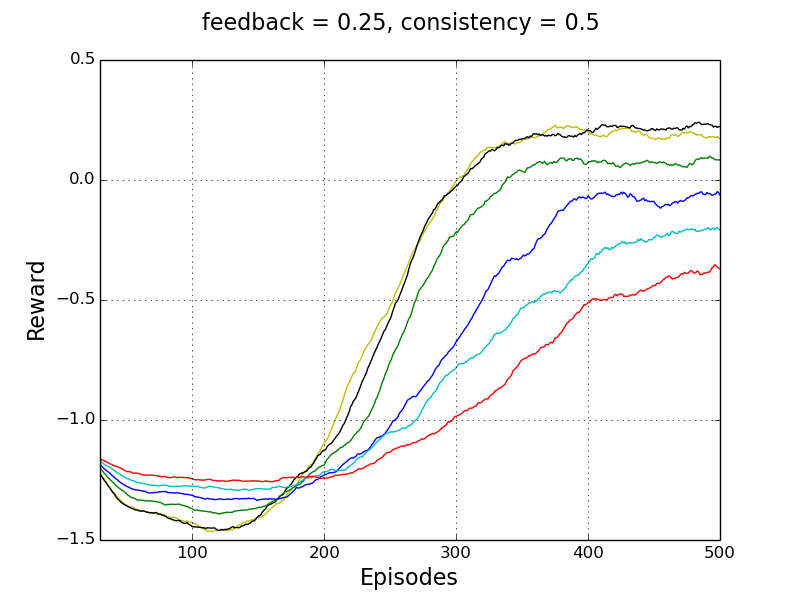}}
\\
\noindent
\subfloat{\includegraphics[width=0.5\textwidth]{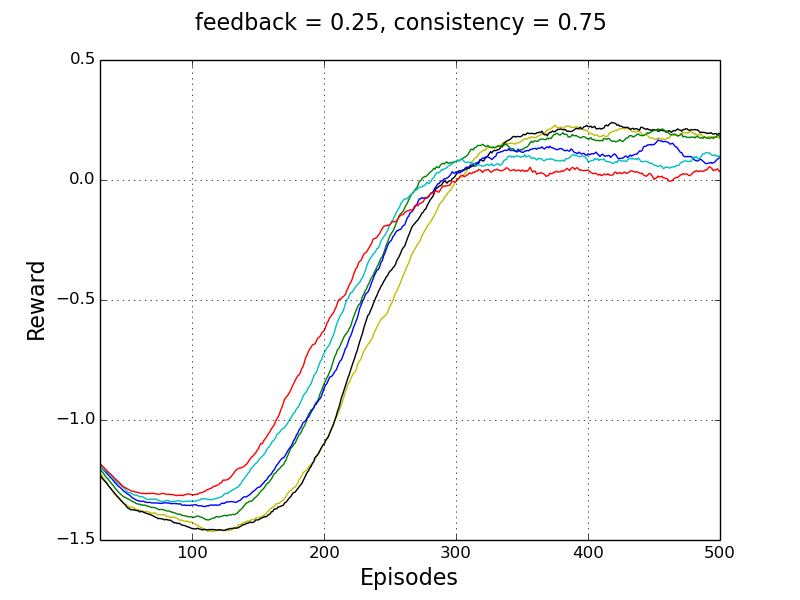}}
\subfloat{\includegraphics[width=0.5\textwidth]{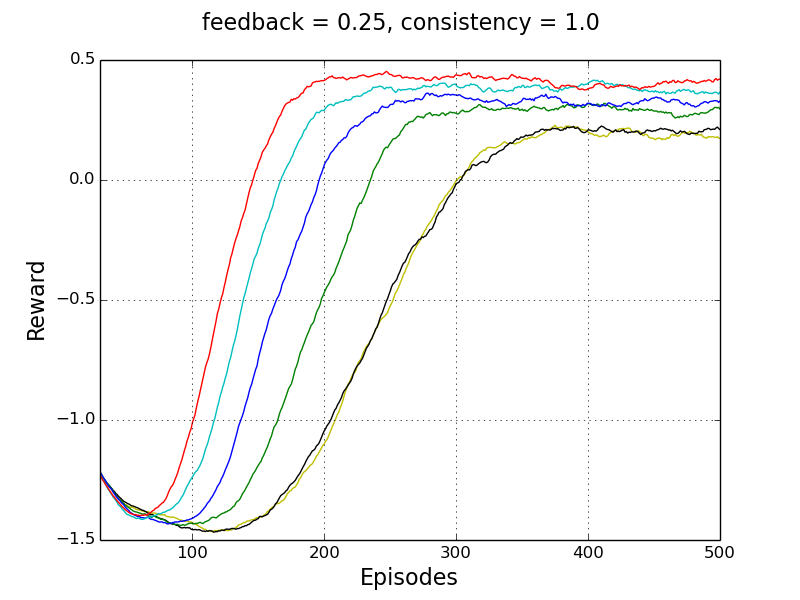}}
\\
\subfloat{\includegraphics[width=1.0\textwidth]{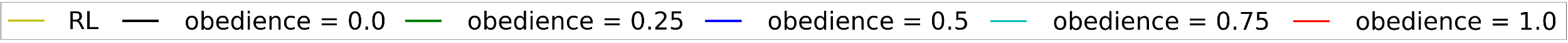}}
\caption[caption]{Collected reward for different values of learner obedience using fixed probability of feedback of 0.25 and four different values for consistency of feedback between $0.25$ and $1.0$.}
\label{fig:rewardf025}
\end{figure}

Initially, we used a fixed probability of feedback $\mathcal{L} = 0.25$, with different values of consistency. 
A similar probability of feedback has been used in (Cruz et al., 2016) and therefore, we used it as a base to start the evaluation. 
The idea then was to test the system over a number of different values of consistency of feedback and learner obedience.
Fig. \ref{fig:rewardf025} shows the collected reward during $500$ episodes for the different values of consistency of feedback $\mathcal{C} \in \left\lbrace0.25, 0.5, 0.75, 1.0\right\rbrace$ and learner obedience $\mathcal{O} \in \left\lbrace0.0, 0.25, 0.5, 0.75, 1.0\right\rbrace$. 
In all cases, the learner obedience $\mathcal{O} = 0$, shown in black, corresponds to autonomous RL which is shown in yellow. 
The collected rewards indicate generally that the more consistent the feedback, the better is the performance. 
Even though that difference in the performance seems to be intuitive, it is important to note that, even with comparatively high values of consistency like $\mathcal{C} = 0.75$, the learner does not achieve significantly better performance compared to autonomous RL while on the other hand, an idealistic perfect consistency ($\mathcal{C} = 1$) allows the learner-agent to achieve much higher collected rewards than with autonomous RL even when the learner obedience is as low as $\mathcal{O} = 0.25$. 
Therefore, in the current scenario, wrong advice has an important negative effect since it does not only lead to the execution of more intermediate steps but also, in many cases, leads to failed-states and thus to a high negative reward ($-1$) and the start of a new learning episode.
Further on in this section, we are going to test additional values of consistency $\mathcal{C} \in [0.75, 1.0]$ to observe how influential small variations in this parameter are.

In Fig. \ref{fig:rewardf025}, agents which follow the advice only $25\%$ of the time ($\mathcal{O}=0.25$), depicted in green, show much better performance when the consistency of feedback $\mathcal{C}$ is lower which is due to the agent being able to ignore the suggested wrong advice and select an action on its own. 
On the contrary, agents which follow the advice all the time ($\mathcal{O}=1.0$), depicted in red color, show much better performance in presence of consistent feedback.

\begin{figure}
\centering
\makebox[\linewidth][c]{%
\subfloat{\includegraphics[width=0.4\textwidth]{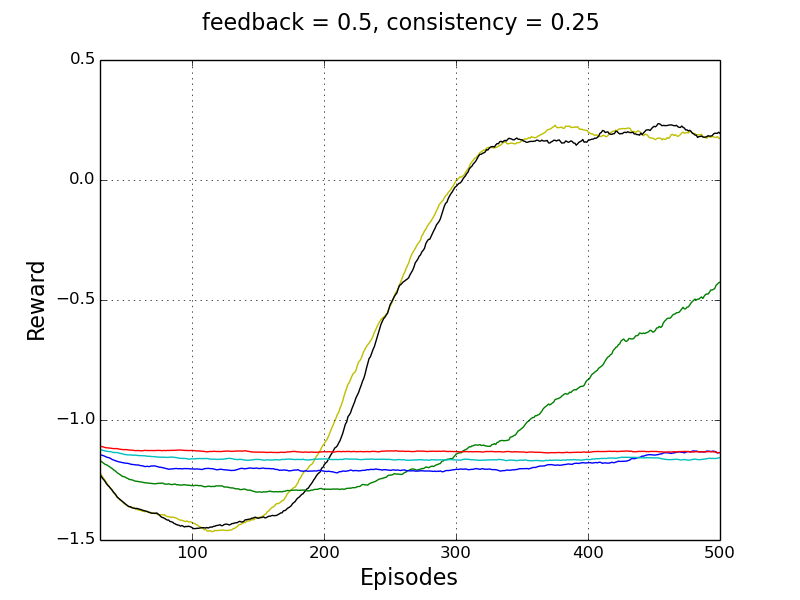}}
\subfloat{\includegraphics[width=0.4\textwidth]{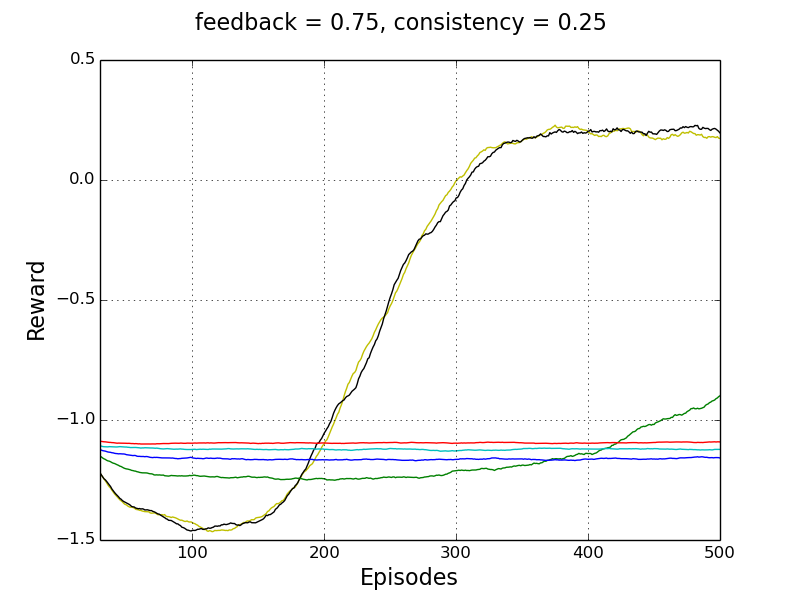}}
\subfloat{\includegraphics[width=0.4\textwidth]{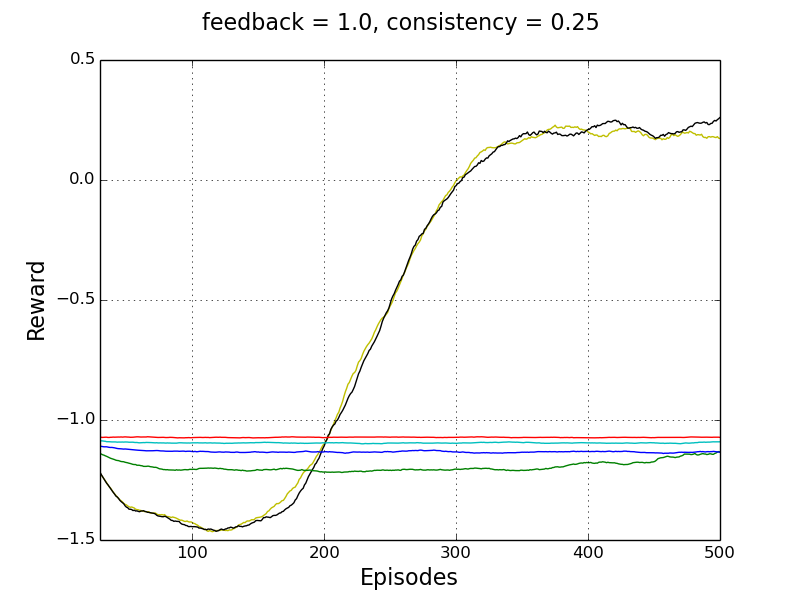}}%
}\\

\makebox[\linewidth][c]{%
\subfloat{\includegraphics[width=0.4\textwidth]{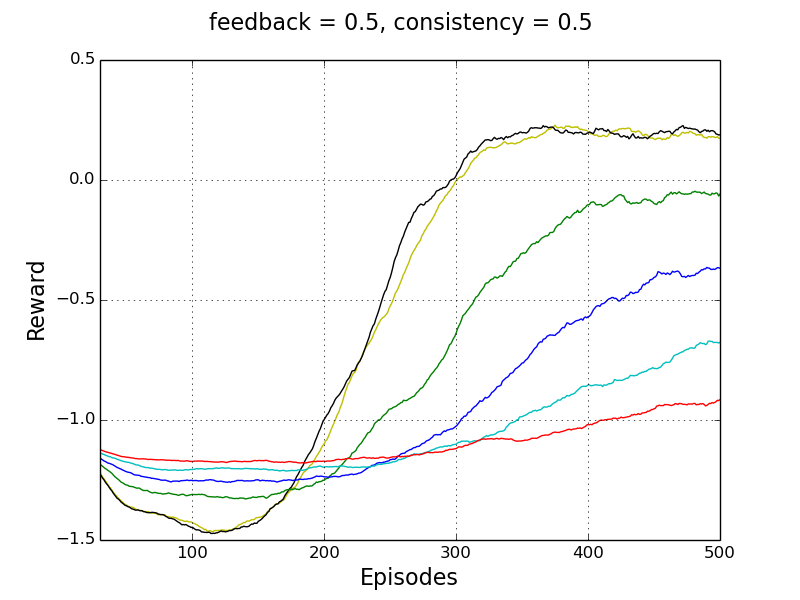}}
\subfloat{\includegraphics[width=0.4\textwidth]{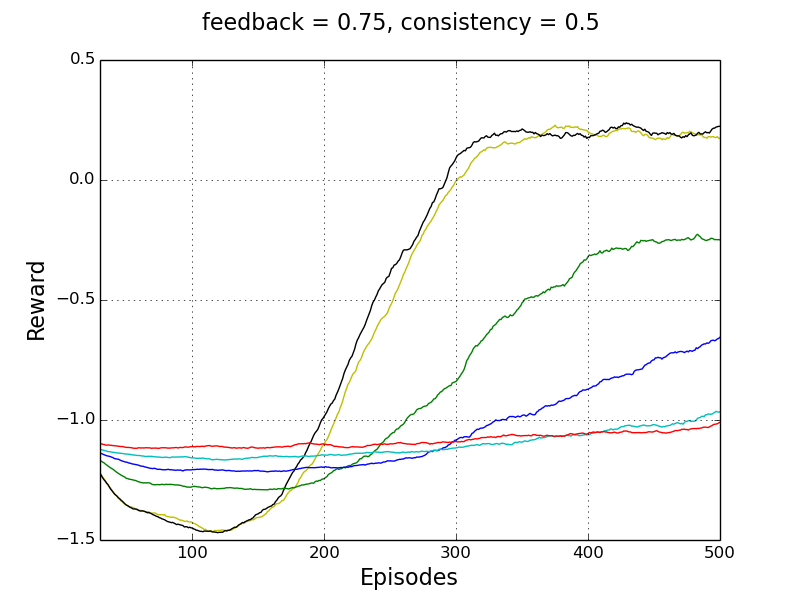}}
\subfloat{\includegraphics[width=0.4\textwidth]{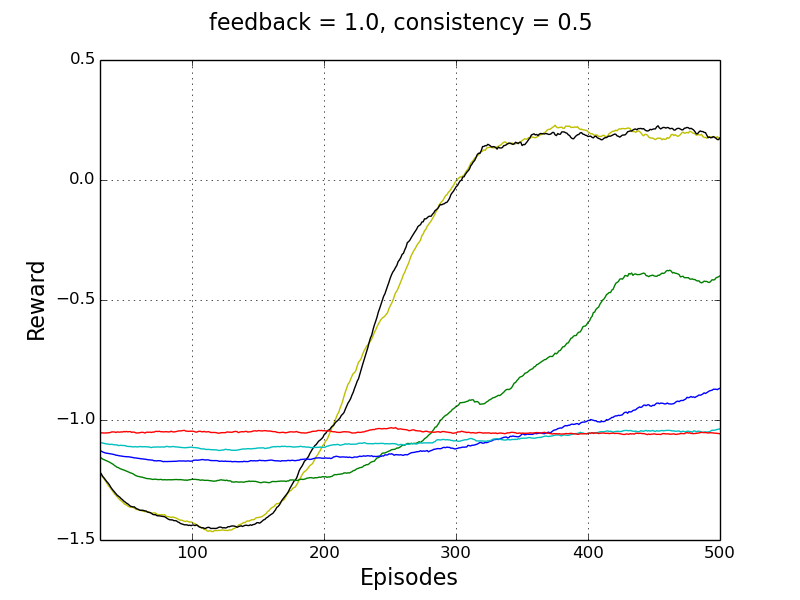}}%
}\\

\makebox[\linewidth][c]{%
\subfloat{\includegraphics[width=0.4\textwidth]{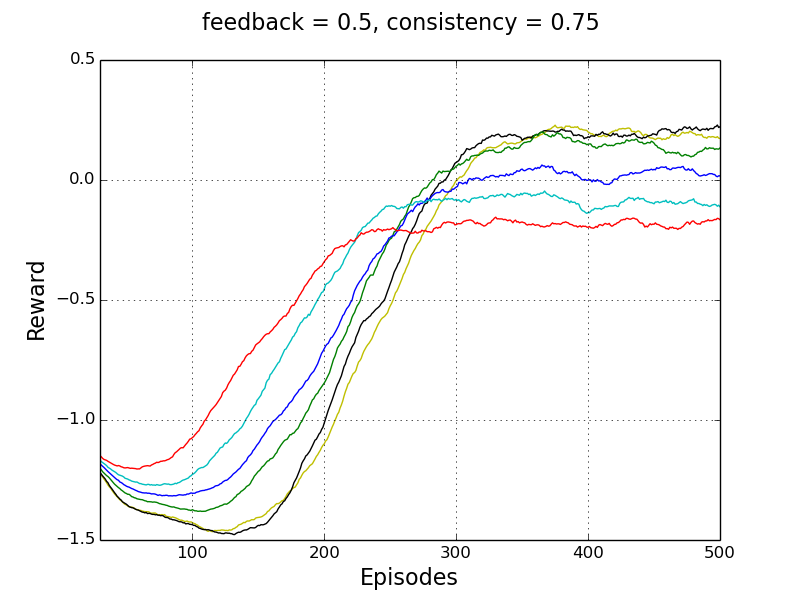}}
\subfloat{\includegraphics[width=0.4\textwidth]{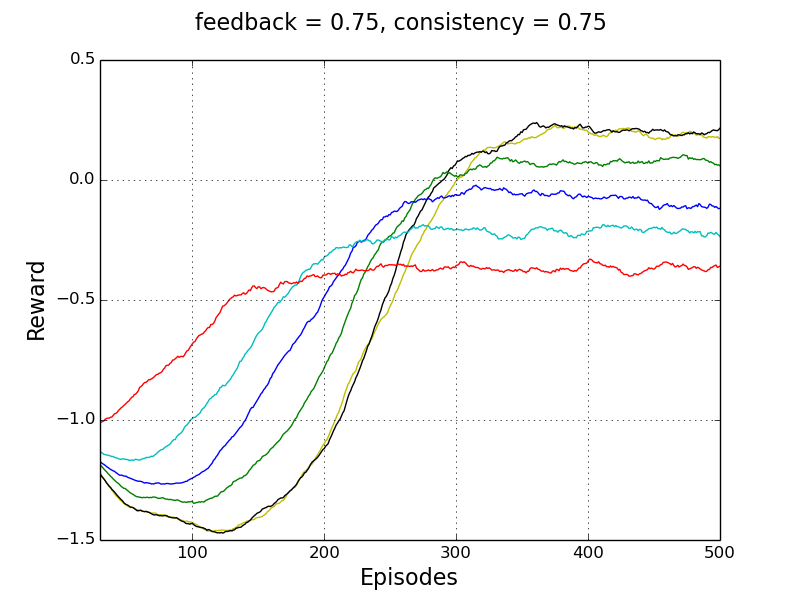}}
\subfloat{\includegraphics[width=0.4\textwidth]{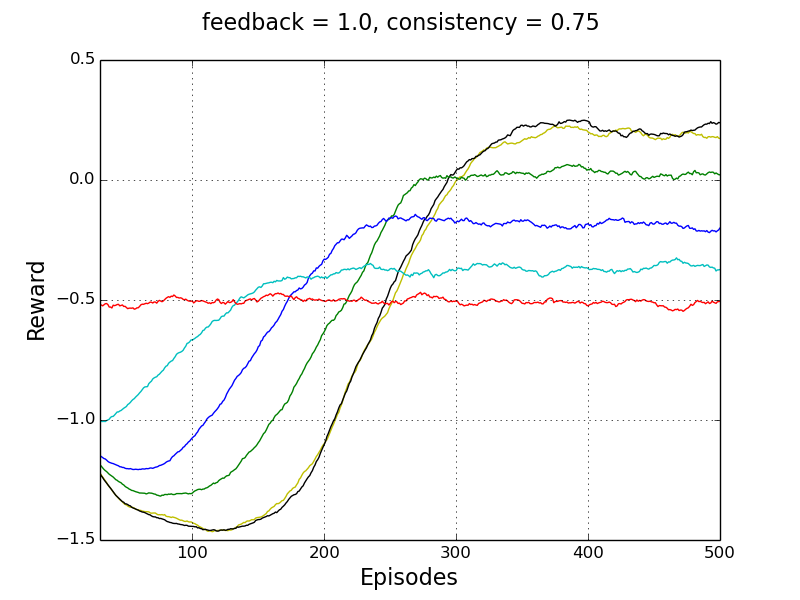}}%
}\\

\makebox[\linewidth][c]{%
\subfloat{\includegraphics[width=0.4\textwidth]{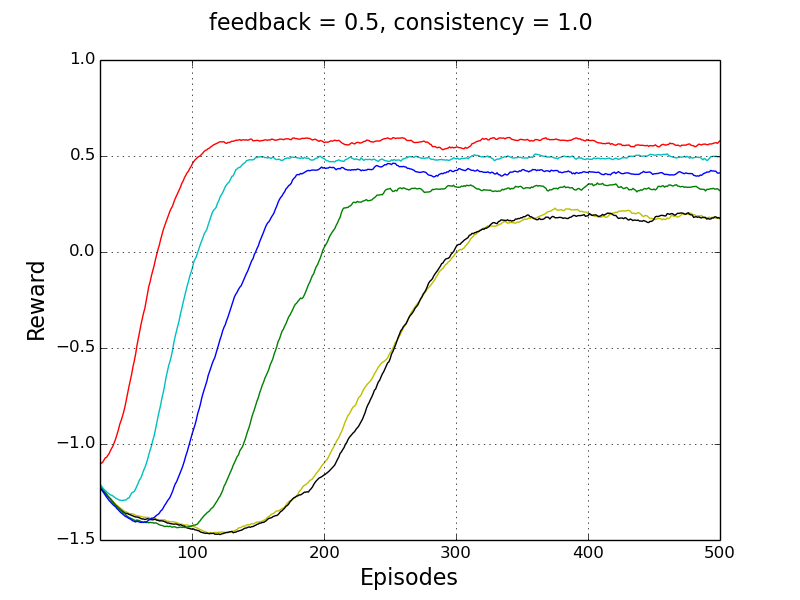}}
\subfloat{\includegraphics[width=0.4\textwidth]{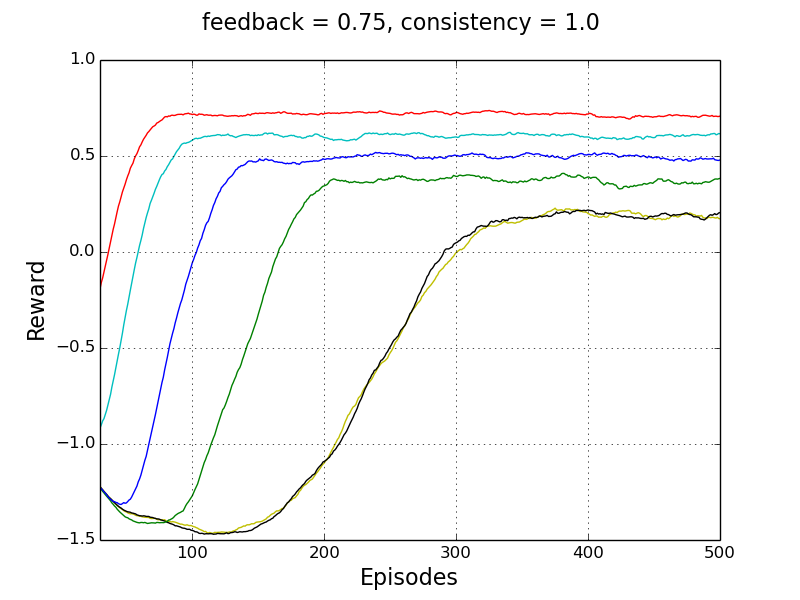}}
\subfloat{\includegraphics[width=0.4\textwidth]{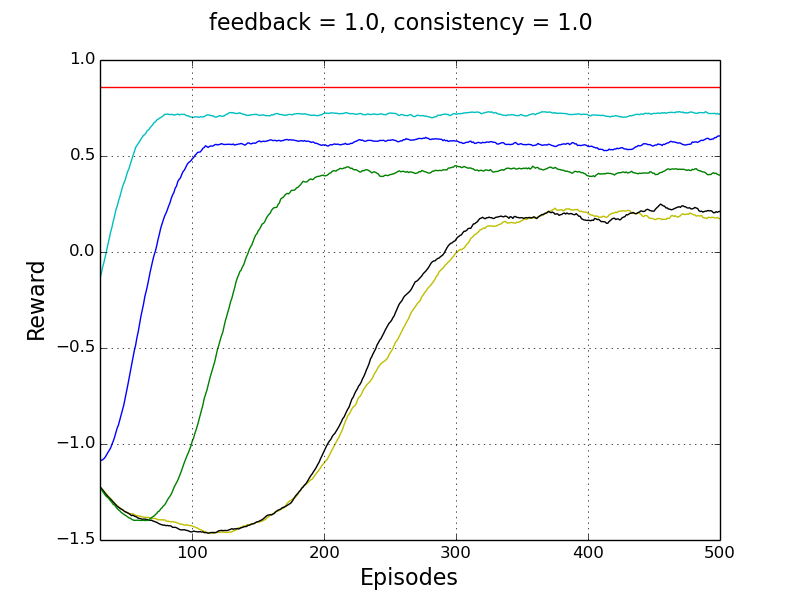}}%
}\\

\subfloat{\includegraphics[width=1.0\textwidth]{figures/label.png}}
\caption[caption]{Collected reward for different learner obedience levels using several probabilities and consistencies of feedback. Higher probabilities of feedback do not necessarily lead to discernible improvements in the overall performance; however, important differences can be noted as higher consistencies of feedback are used.}
\label{fig:rewardsfc}
\end{figure}

Thereupon, we modified the probability of feedback for the purpose of testing how influential different consistencies of feedback $\mathcal{C}$ and different learner obedience levels $\mathcal{O}$ are. 
Fig. \ref{fig:rewardsfc} shows the accumulated reward during $500$ episodes for probability of feedback $\mathcal{L} \in \left\lbrace0.5, 0.75, 1.0\right\rbrace$ (the outcome using probability of feedback of $0.25$ is already shown in Fig. \ref{fig:rewardf025}) and consistency of feedback $\mathcal{C} \in \left\lbrace0.25, 0.5, 0.75, 1.0\right\rbrace$ using learner obedience $\mathcal{O} \in \left\lbrace0.0, 0.25, 0.5, 0.75, 1.0\right\rbrace$.

In Fig. \ref{fig:rewardsfc} the columns show the performance over different probabilities of feedback, while the rows show the performance over different values of consistency.
Observing each row, it can be seen that higher probabilities of feedback do not considerably improve the outcomes in terms of the collected reward, suggesting that often interactive feedback does not necessarily enhance the overall performance but it is rather the consistency of feedback that makes prominent differences.
In fact, observing the outcomes down the columns, thus with the same probability of feedback, different values of consistency lead to significant improvements in the collected reward and consequently, consistency of feedback has much more impact on the final learning performance.
For instance, when using the consistency of feedback $\mathcal{C} = 1.0$ (fourth row in Fig. \ref{fig:rewardsfc}), in all cases the accumulated reward is higher than $0.5$, but on the other hand, when using the consistency of feedback $\mathcal{C} = 0.75$ (third row in Fig. \ref{fig:rewardsfc}), the accumulated reward tends to slightly decrease as trainer advice increases, meaning that more interactive feedback does not help in the presence of poor consistency of feedback or, in other words, of bad advice.

\begin{figure}
\centering
\subfloat{\includegraphics[width=0.5\textwidth]{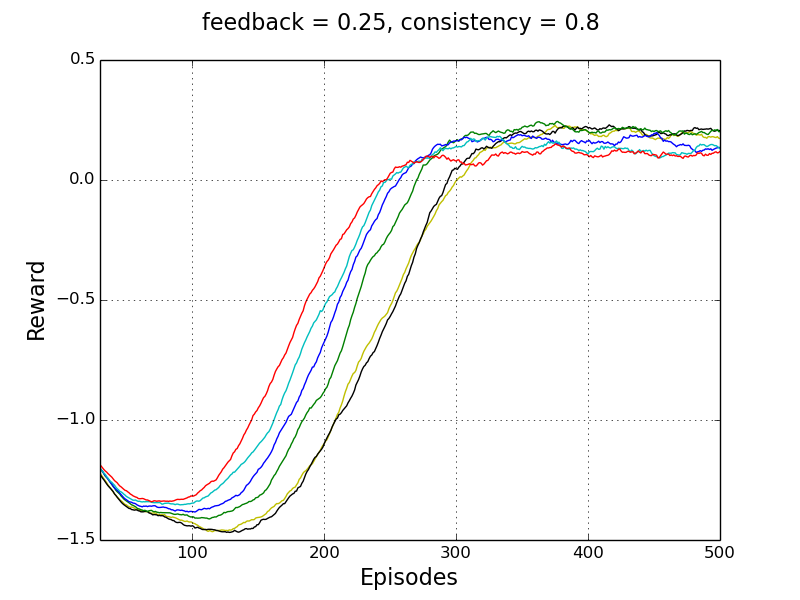}}
\subfloat{\includegraphics[width=0.5\textwidth]{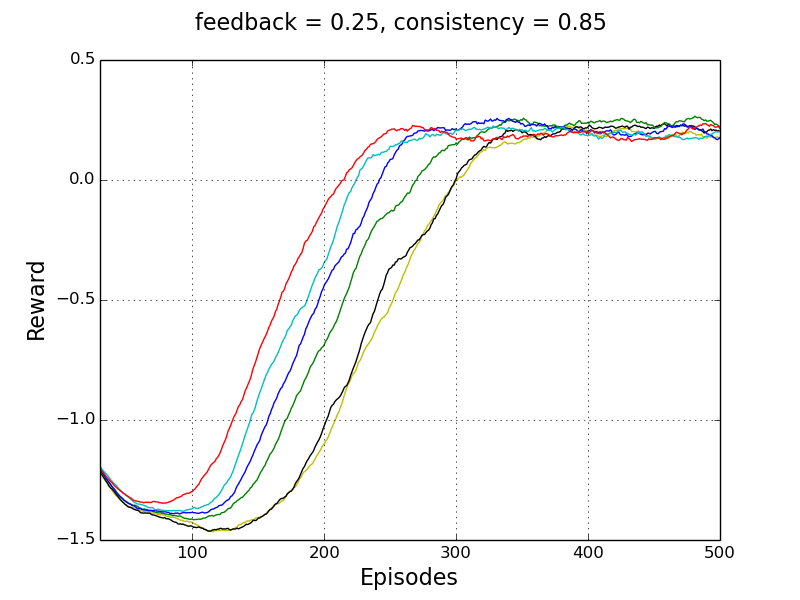}}
\\
\noindent
\subfloat{\includegraphics[width=0.5\textwidth]{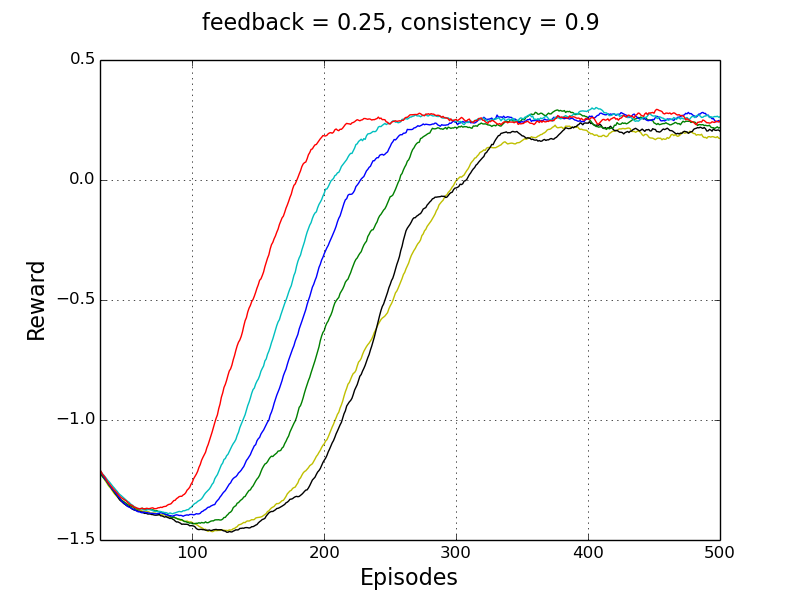}}
\subfloat{\includegraphics[width=0.5\textwidth]{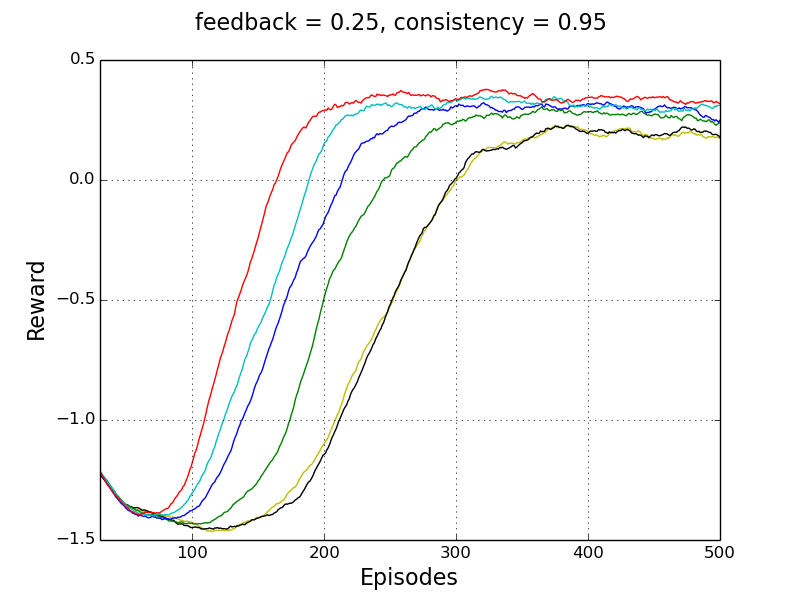}}
\\
\subfloat{\includegraphics[width=1.0\textwidth]{figures/label.png}}
\caption[caption]{Collected reward for different values of learner obedience using fixed probability of feedback $0.25$ and for four different cases for higher consistencies of feedback between $0.8$ and $0.95$.}
\label{fig:rewardsCSmall}
\end{figure}

Ultimately, since the consistency of feedback shows considerable sensibility in the presence of small variations, we performed one additional experiment keeping the probability of feedback fixed to $\mathcal{L} = 0.25$ as in Fig. \ref{fig:rewardf025} since we use this value as a base as aforementioned.
We tested the consistency of feedback with values $\mathcal{C} \in \left\lbrace0.8, 0.85, 0.9, 0.95\right\rbrace$ (consistency of $0.75$ and $1.0$ are already shown in Fig. \ref{fig:rewardf025}) to evaluate how these slight changes impact on the overall performance.
Fig \ref{fig:rewardsCSmall} shows the accumulated rewards for learner obedience $\mathcal{O} \in \left\lbrace0.0, 0.25, 0.5, 0.75, 1.0\right\rbrace$. 
It can be seen that such small differences in the consistency of feedback can lead to dissimilar outcomes, ranging from behavior similar to autonomous RL when $\mathcal{C} = 0.8$ to behavior similar to a fully and correctly advised learner-agent when $\mathcal{C} = 0.95$. 
Therefore, even a small proportion of bad advice can considerably impoverish the learning process, which shows how important it is to select trainers that can give useful advice in most states since specialised trainers, despite being more successful themselves from the initial state, have limited knowledge when it comes to states that lie outside their specialised policy. 

In our approach, we have used the probability of feedback as a way to control how much advice is given to the learner-agent in terms of assistance during selected training episodes. 
As mentioned above, the consistency of feedback allows to mimic the behavior of human trainer-agents who are susceptible to make mistakes during the learning process.
Nevertheless, at this point, all the instances of advice are received by the learner-agent without any discrimination between right or wrong advice. 
As discussed, the inconsistent feedback may in fact lead to slow the learning process in terms of accumulated reward. 
Therefore, the learner obedience parameter is an effective way for learner-agents to suppress the influence of the inconsistent feedback disregarding some wrong pieces of advice. 
In this way, the learner-agents are able to accumulate more reward during the learning process.

\section{Conclusions and Future Work}

In this work, we presented a comparison of artificial agents that are used as parent-like teachers in an IRL cleaning scenario.
We have defined three classes of trainer-agents related to our scenario. 
The agents differ in their characteristics and consequently in the obtained performance during their own learning process and in turn as trainers. 
The three agents vary in their main properties which reflect in their behavior as i) the specialist-A agent with the largest accumulated reward, ii) the specialist-B agent with the largest amount of experience in terms of the number of explored states, and iii) the polymath agent with the smallest standard deviation.

It has been shown that there exists divergence in the internal representation of the knowledge of the agents through state--action Q-values since there are states in which it is not possible to distinguish what actions lead to greater reward. 
Using the polymath agent as an advisor leads to both greater reward and faster convergence of the reward signal and also to a more stable behavior in terms of the state visit frequency of the learner-agents, which can be seen in the standard deviation for each visited state when compared with the case of the specialist-A agent as a trainer.

IRL generally helps to improve the performance of an RL agent using parent-like advice. 
Nonetheless, it is important to take into account that higher levels of interaction do not necessarily have a direct impact on the total accumulated reward. 
More importantly, the consistency of feedback seems to be more relevant when dealing with different learner obedience parameters (or a noisy or unreliable communication channel) since small variations can lead to considerably different amounts of collected reward.

Agents with a smaller standard deviation are preferred candidates to be parent-like teachers since they have a much better distribution of knowledge among the states. 
This allows them to adequately advise learner-agents on what action to perform in specific states. 
Agents with biased knowledge distributions collect more reward themselves, but nevertheless, have a subset of states where they cannot properly advise learners.
This leads to a worse performance in the apprenticeship process in terms of maximal collected reward, convergence speed, and behavior stability represented as the standard deviation for each visited state.

The finding that an expert in a certain domain is not necessarily a good teacher might also help the understanding of biological or natural systems in terms of assistive teaching.
For instance, a good soccer player is not necessarily a good soccer trainer.
We are not aware of studies that confirm this in biological systems or human-human interaction. 
However, \cite{Taylor11} gave some interesting insights about a human-agent interaction approach.
Also, \cite{Griffiths12} studied different teacher behaviors to improve the apprenticeship in learner-agents.
Although their experiments are based on human-human interaction, they have used tutors that have mastered a given task without any classification about the level of expertise.

An important future work is to investigate how the obtained results can be scaled up to either larger discrete or continuous scenarios. 
There are many real-world problems which have inherently continuous characteristics.
Many of these problems have been addressed using autonomous RL by discretizing the state-action space. 
This discretization may lead to the introduction of hidden states or hidden actions for the RL agent. 
However, a human trainer may not know or have access to this discrete representation and may advise actions which are not directly mapped into the discrete action-state representation used by the learner-agent.
Therefore, if the learner-agent maps the given advice into the discrete representation, it could lead to a slight error which over time could be accumulated rendering the learned policy useless.
An alternative is to address the problem directly in its continuous representation, but to the best of our knowledge, continuous IRL has not been studied yet.
It can be expected that RL agents have similar behavior in continuous scenarios compared to discrete ones since they are designed to find the optimal solution maximizing the collected reward.

Moreover, adaptive learner behavior can be explored, thus allowing to decide which advice to follow depending on the collected knowledge about the current state that the learner-agent has at a specific time.
Then, the learner-agent would act with diverse values for the learner obedience parameter, adapting it in real time. 
Greater learner obedience can be expected at the beginning of the learning process, but over time the learner-agent should take its own experience more into account and therefore follow its own policy instead of the parent-like advice, leading to smaller obedience values. 
In the same way, if new space is explored and consequently the reward gets worse, then parent-like advice could be used once again, leading to a dynamic learning process, taking advice into account when necessary while avoiding bad advice when possible.

\section*{Acknowledgements}
The authors gratefully acknowledge partial support by CONICYT scholarship 5043, the German Research Foundation DFG under project CML (TRR 169), the European Union under project SECURE (No 642667), and the Hamburg Landesforschungsf\"orderungsprojekt CROSS.

\end{document}